%% file: main.tex
\documentclass[10pt,twocolumn,letterpaper]{article}

\PassOptionsToPackage{dvipsnames,svgnames,x11names}{xcolor}

\usepackage[pagenumbers]{iccv}
\usepackage{times}
\usepackage{epsfig}
\usepackage{graphicx}
\usepackage{amsmath}
\usepackage{amssymb}
\usepackage{multirow}
\usepackage{caption}
\usepackage{etoc}
\usepackage{minitoc}

\usepackage[pagebackref=true,breaklinks=true,letterpaper=true,colorlinks,bookmarks=false]{hyperref}

\definecolor{cvprblue}{rgb}{0.21,0.49,0.74}
\definecolor{lightblue}{HTML}{00B0F0}
\definecolor{svcolor}{HTML}{E97132}
\PassOptionsToPackage{pagebackref,breaklinks,colorlinks,allcolors=cvprblue}{hyperref}
\newcommand{\ourwork}{Dress\&Dance\xspace}
\newcommand{\fancyname}{{\bf\textcolor{lightblue}{Dress}\&\textcolor{red}{Dance}}\xspace}
\newcommand\themodel{\ourwork} 
\def\sv{{\href{https://immortalco.github.io/DressAndDance/}{project page}}\xspace}
\newcommand{\svat}[1]{\sv}

\newcommand{\tabdot}{Tab\onedot}
\newcommand{\figdot}{Fig\onedot}

\newcommand{\suba}{\textcolor{red}{(a)}\xspace}
\newcommand{\subb}{\textcolor{red}{(b)}\xspace}
\newcommand{\subc}{\textcolor{red}{(c)}\xspace}

\begin{document}

\title{\fancyname: Dress up and Dance as You Like It\\Technical Preview}

\author{Jun-Kun Chen$^{1}$ \qquad Aayush Bansal$^{2}$ \qquad Minh Phuoc Vo$^{2}$\qquad Yu-Xiong Wang$^{1}$ \vspace{0.1em} \\ 
    $^1$University of Illinois Urbana-Champaign \qquad $^2$SpreeAI\vspace{0.1em}\\
    {\tt \hspace{0mm}\{junkun3,yxw\}@illinois.edu \qquad \tt \{aayush.bansal,minh.vo\}@spreeai.com} 
    \vspace{0.2em}\\
    {\tt \href{https://immortalco.github.io/DressAndDance/}{immortalco.github.io/DressAndDance}}
}

\input{sec/0_abstract}

\input{sec/1_intro}

\input{sec/2_related}
\input{sec/3_method_simple}
\input{sec/4_expr_v2}

\input{sec/5_conclusion}

{\small
\bibliographystyle{ieee_fullname}
\bibliography{egbib}
}

\end{document}

%% file: sec/0_abstract.tex
\twocolumn[{
\maketitle
\vspace{-5mm}
\renewcommand\twocolumn[1][]{#1}
    \centering
    
    \includegraphics[width=\textwidth]{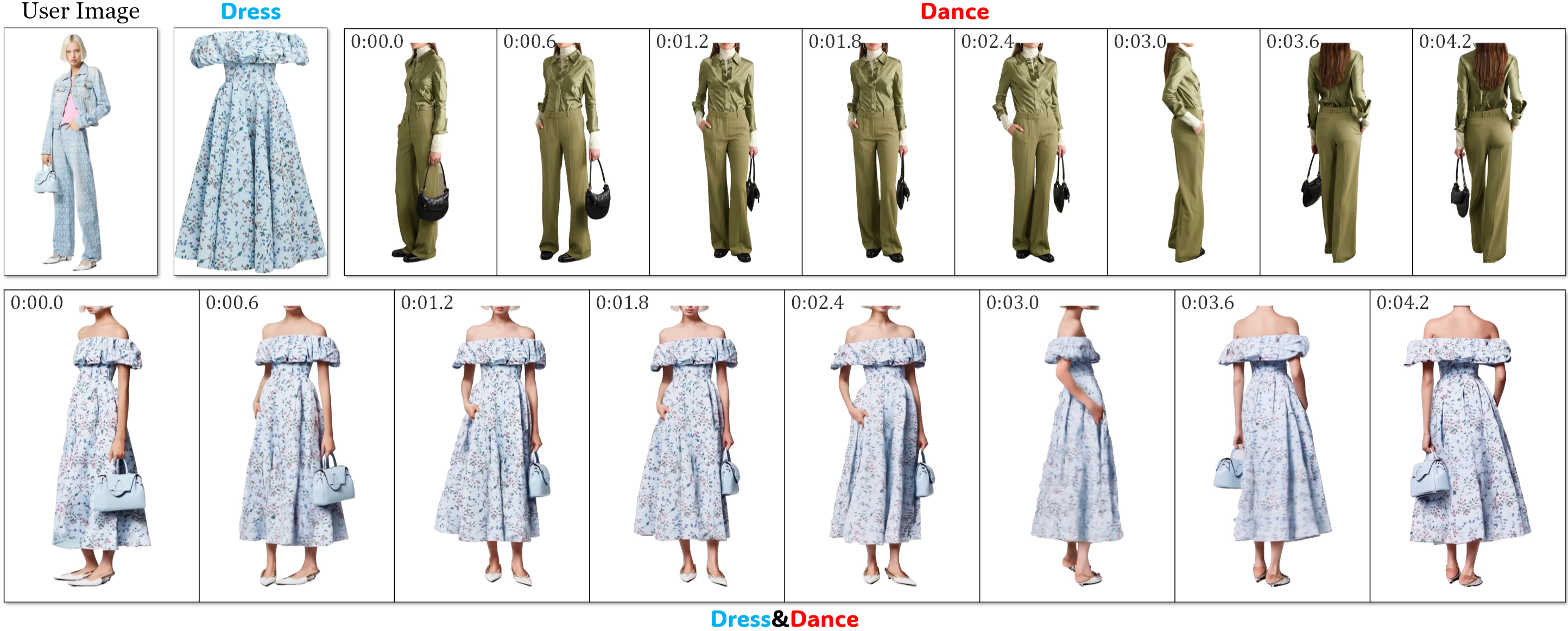}
    \vspace{-6mm}
    \captionof{figure}{Given a single image of a user, a garment (\textcolor{lightblue}{\emph{dress}}) that they would like to wear, and an example video showing how they would like to animate themselves (\textcolor{red}{\emph{dance}}) as shown in the first row. Our method, \fancyname, generates a high-quality $5s$ video ($1152\times720$, $24$ FPS) of the user wearing the target garment containing desired motion while maintaining their accessories such as bag and shoes.}

    \label{fig:teaser}

    \vspace{2mm}

}]

\begin{abstract}
We present \fancyname, a video diffusion framework that generates high quality
$5$-second-long $24$~FPS virtual try-on videos at $1152\times720$ resolution of a user wearing desired garments while moving in accordance with a given reference video. Our approach requires a single user image and supports a range of tops, bottoms, and one-piece garments, as well as simultaneous tops and bottoms try-on in a single pass. Key to our framework is CondNet, a novel conditioning network that leverages attention to unify multi-modal inputs (text, images, and videos), thereby enhancing garment registration and motion fidelity. CondNet is trained on heterogeneous training data, combining limited video data and a larger, more readily available image dataset, in a multi-stage progressive manner. Dress\&Dance outperforms existing open source and commercial solutions and enables a high quality and flexible try-on experience.  
\end{abstract}

%% file: sec/1_intro.tex
\section{Introduction}

\begin{figure*}
  \includegraphics[width=\textwidth]{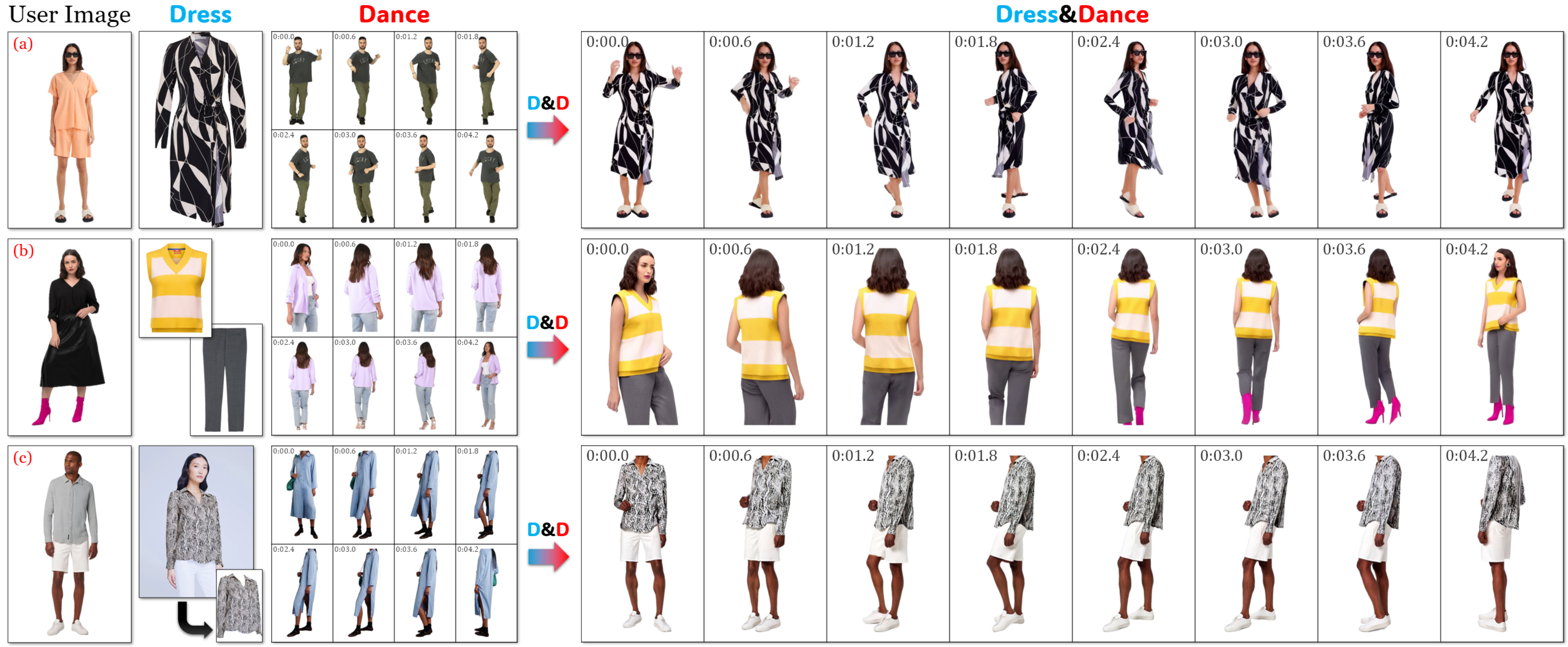}
  \vspace{-7mm}
  \caption{\fancyname can be used in multiple ways:  \suba \themodel can generate complicated dancing motions; \subb \themodel performs virtual try-on for both top and bottom garments simultaneously; and \subc \themodel can input garments worn by other users. }
  \label{fig:modes}
  \vspace{-6.5mm}
\end{figure*}

\label{sec:intro}
Imagine yourself trying a new dress in a fitting room. It is natural to move around to get a feel for how the garment fits and flows. Current computational methods that enable virtual garment try-on typically generate static 2D images. Viewing oneself in a single image does not provide the full and expressive experience of a try-on. In this paper, we introduce \fancyname, a video diffusion framework that generates a $5$-second, $24$~FPS video at $1152 \times 720$ resolution of a user wearing a desired garment. The motion in the generated video is guided by an example video selected by the user. Our approach allows the user to choose the garment ({\bf\color{lightblue} Dress}) and control how they would like to animate ({\bf\color{red} Dance}) with it, as illustrated in Fig.~\ref{fig:teaser}. \themodel supports a wide range of tops, bottoms, and one-piece garments -- as well as their combinations at once. 
Our model allows for try-ons of clothing borrowed from another human and garments and accessories modification using text prompts. Fig.~\ref{fig:modes} highlights a few applications of \themodel.

\begin{figure*}
\centering
  \includegraphics[width=1\textwidth]{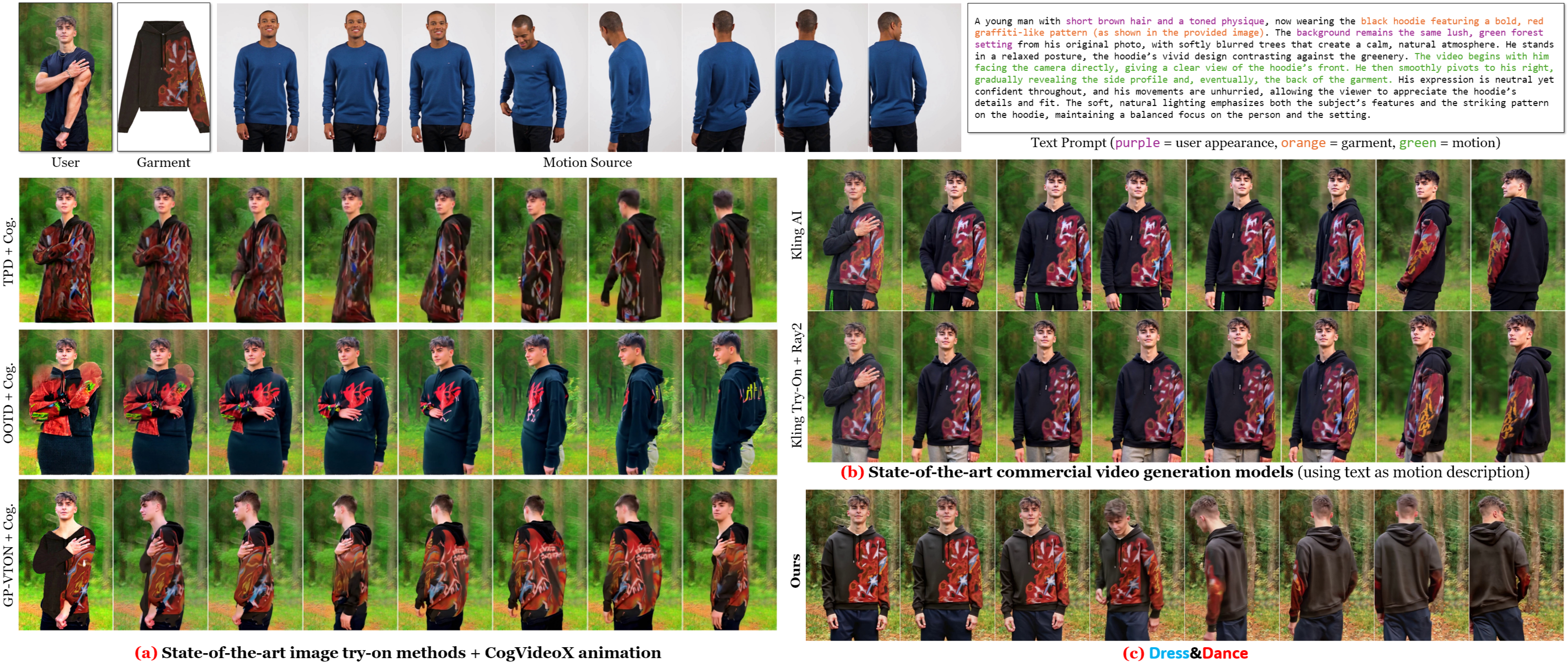}
  \vspace{-7mm}
  \caption{Given a user image, a desired garment, and an indicated reference video, we extract a detailed text description via GPT \cite{gpt} as shown in the first row. 
  \suba Single image try-on methods TPD~\cite{TPD}, OOTD~\cite{ootdiffusion}, and GP-VTON~\cite{gpvton} struggle to generate meaningful try-on, whilst error propagation happens when motion is applied via CogVideoX~\cite{cogvideox}.
  \subb State-of-the-art commercial models, such as Kling~\cite{kling}, are able to perform try-on but still struggle in capturing the nuanced motion description, as \emph{text} alone is not sufficient for motion description. This is also evident when we use Ray2~\cite{ray2} for motion generation even when using Kling for try-on.  
  Also, as the user's right hand covers part of the garment, the information of the covered patterns is lost for the video generation model, leading to incorrect garment appearances when the hand moves in the video, for all baselines in \suba and \subb.
  \subc Our \fancyname generates high-quality virtual try-on results that faithfully preserve both garment and user appearance with precise motion, even when the hand moves off.
}
  
  \label{fig:expr:gallery:singlegar}
  \vspace{-6.5mm}
\end{figure*}

The emergence of diffusion models~\cite{sd} has allowed the generation of videos from text~\cite{vidrd,cogvideox} and images~\cite{svd,i2vgen,kling}. Generating videos from text requires highly detailed descriptions; otherwise, the model struggles to capture essential details. In our context, it is difficult to describe the content textually such that both the user's appearance and the integrity of the target garment are preserved in the generated results. To sidestep the problem, one strategy is to first perform a single image try-on~\cite{TPD,ootdiffusion,gpvton,hrvton} and then animate the output with video diffusion models like CogVideoX~\cite{cogvideox}. We observe temporally incoherent outputs due to error propagation from the first frame (\figdot\ref{fig:expr:gallery:singlegar}-\suba), particularly when there are significant pose changes, leading to occlusions of the garment or human body. Another challenge is to instruct the machine learning model to generate desired motion. Simple movements, such as \textit{turn-around}, can be described textually. Capturing more nuanced movements is difficult with text, even when using state-of-the-art commercial methods such as Kling~\cite{kling} and Ray2~\cite{ray2}, as shown in \figdot\ref{fig:expr:gallery:singlegar}-\subb. We overcome these challenges by learning an end-to-end approach that leverages exemplar videos to describe nuanced motion. 
Our results faithfully preserve garment and user appearance details and strictly follow the direction, as shown in \figdot\ref{fig:expr:gallery:singlegar}-\subc.

The key to our work is a conditioning network, CondNet, that leverages cross-attention for conditioning on multiple modalities (text, images, and videos). By converting \emph{heterogeneous} conditions into \emph{homogeneous} attention sequences, we can concatenate them with the sequences of video and text for cross-attention, enabling the model to handle them in a unified manner. CondNet implicitly connects all pairs of the input image's pixels to the generated video's pixels through cross-attention. This allows us to better transfer garment details, supports multiple garment types, including garment combination simultaneously, and is robust to varying garment capture methods. Whether the garment is casually captured as a flat image or worn by an individual, our model can effectively handle both scenarios.

Due to the high compute cost of attention modules in generating high-resolution visual output, we design a stage-wise, multi-phase, resolution progressive training strategy to effectively and efficiently train the model. First, we adopt a curriculum garment warm-up learning to guide the coarse garment try-on location on the user body and a progressive resolution training to preserve the user and garment identities. Secondly, we further boost the resolution and quality, by introducing an auto-regressive video refiner stage to upsample the video from $8$~FPS to $24$~FPS while further refining the appearance details.  
We show ablation results that demonstrate the effectiveness of this training strategy.

\textbf{Our contributions:}  (1) We propose CondNets, a simple yet effective conditioning strategy that unifies multi-modal inputs through cross-attention, improving both garment registration and overall try-on quality. (2) We introduce compute-effective and data-efficient training strategies that allow us to generate high-resolution outputs with limited compute and data. (3) We show experimentally that our simple method can effectively handle arbitrary garments and is robust to various garment capture methods, enabling the user to simultaneously try on a set of garments or point to a garment worn by another person and perform a try-on. Our method achieves state-of-the-art quality in various challenging scenarios, outperforming open source baselines and commercial models such as Kling \cite{kling} and Ray2 \cite{ray2}.

%% file: sec/2_related.tex
\section{Related Work}
\label{sec:related}

\noindent\textbf{Videos from a Single Image.} As a representative method, Stable Video Diffusion (SVD) \cite{svd} generates short videos from a single image. However, it supports only landscape videos and is limited to short video lengths. Similarly, I2VGen-XL \cite{i2vgen} uses image and text input to generate short videos, but is restricted to landscape formats. A more flexible solution, CogVideoX-I2V \cite{cogvideox}, supports portrait videos and allows image-to-video synthesis with text guidance, providing more control over generated content. Commercial models, such as Kling Video 1.6 \cite{kling} and Ray2 \cite{ray2}, offer a similar capability for image-to-video synthesis, supporting portrait videos with image and text inputs, enabling a more customizable output. These methods focus on generating video sequences from still images but do not address garment try-on specifically or the need for motion-guided generation. Note that Kling has a standalone image try-on model which we use for analysis in this work.

\begin{figure*}
\centering
  \includegraphics[width=1\textwidth]{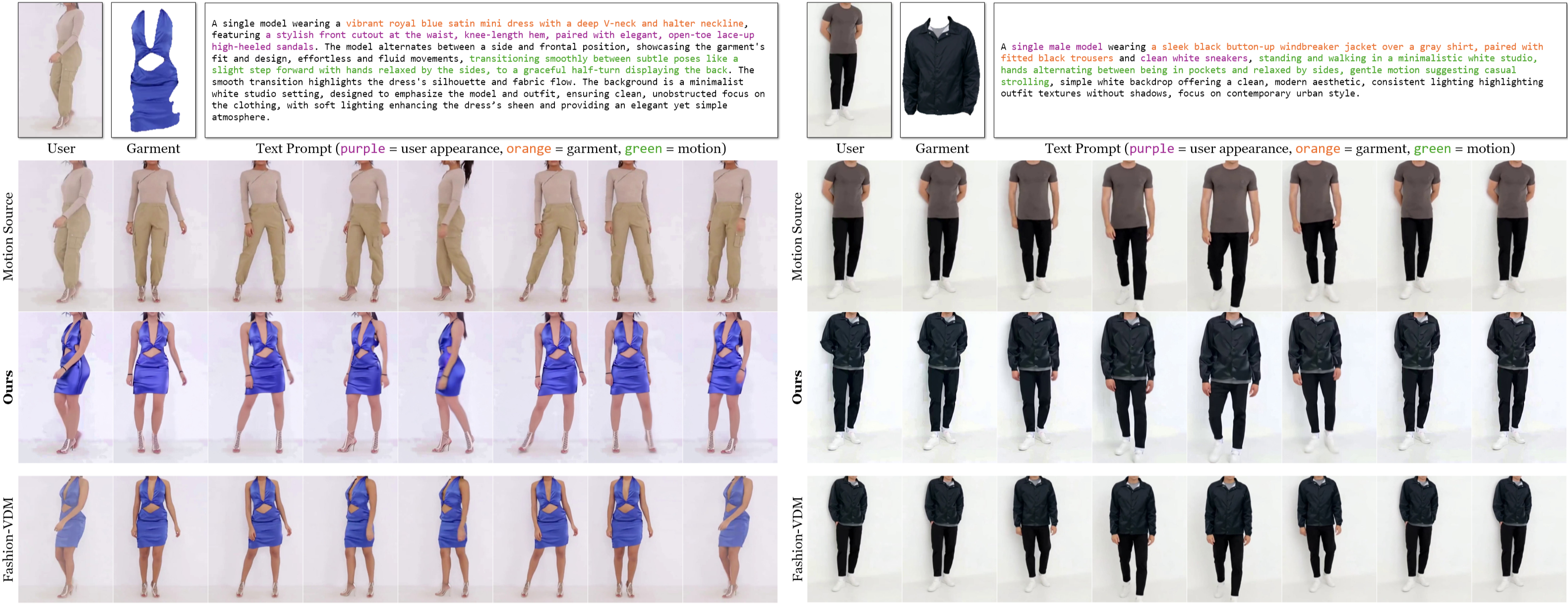}
  \vspace{-7mm}
  \caption{Our \fancyname supports transferring a garment from another given image via segmentation, regardless of the pose of that image. Notably, our \fancyname generates high-resolution videos at $1152\times 720$ with clear appearance and more details, while the baseline Fashion-VDM \cite{fashionvdm} exhibits color fading and generates low-resolution $512\times 384$ videos.}
  \label{fig:expr:gallery:seggar}
  \vspace{-6.5mm}
\end{figure*}

\noindent\textbf{Single 2D Image Try-On.} This is the most popular virtual try-on work stream and supports both one or multiple target garment(s) to produce a single try-on image as output~\cite{stablegarment,stablevition,TPD,ootdiffusion,gpvton, hrvton, 4dvton,tryondiffusion, pfafn, sizematters,mmvto,tryonadaptor}. These methods typically require \emph{paired} training data consisting of the garment and corresponding human model images wearing that garment. To prevent information leakage from such paired training data, they extract garment agnostics human intermediates, such as the pose or try-on region mask, from the human model images, and train the model to resynthesize the original RGB human model images in a self-supervised setting. However, at inference time, the model is given the \emph{unpaired} setting where it needs to try-on a chosen garment on a user image wearing a different garment, making inference harder than training due to information leakage at training stage. Moreover, the use of the intermediate representations also propagate their pre-processing artifacts to the try-on results. Our method sidesteps both problems using synthetically generated unpaired triplets to not only eliminate the need for these intermediates, but also to enable a \emph{consistent} training setting as evaluation, potentially improving the try-on results in the evaluation.

\noindent\textbf{Video-to-Video Translation.} Video-to-video translation methods have explored both supervised and unsupervised approaches to generate temporally consistent videos. VideoShop~\cite{videoshop} is a training-free approach that relies on a pre-trained diffusion model to edit the first frame of a video, which is then propagated across subsequent frames to maintain temporal consistency. Similarly, BIVDiff \cite{bivdiff} performs per-frame editing, refining the video using a pre-trained diffusion model to ensure consistency across frames. Another relevant approach is CogVideoX-V2V \cite{cogvideox}, employing a pre-trained text-to-video diffusion model combined with SDEdit \cite{sdedit} for text-guided video editing. 
These methods focus on video editing and manipulation, but do not address the specific needs of garment try-on or the combination of motion and garment generation. \themodel seamlessly integrates garment try-on within video generation, allowing users to choose a desired garment and manipulate motion, effectively combining the adaptability of video-to-video translation with the added functionality of virtual try-on.

\noindent\textbf{Video Virtual Try-On (VVT).} Early work of VVT \cite{fwgan,mvton,clothformer,shineon} utilizes generative adversarial networks (GANs) \cite{gan} to perform virtual try-on, and warp previous frames to new frames to form videos. After the emergence of diffusion models \cite{svd}, follow-up work \cite{tunnel,gpdvvto,vivid,wildvidfit,fashionvdm} leverages image or video diffusion models for VVT. However, these methods produce low-resolution (lower than 512$\times$512) VVT videos with fewer frames. Additionally, the diffusion-based methods work only on UNet-based diffusion models, which limits their scalability. In comparison, our \themodel generates high 1152$\times$720 resolution, $121$-frame videos, through a powerful DiT-based video diffusion model.

%% file: sec/3_method_simple.tex
\section{\fancyname: Methodology Overview}
\label{sec:method}

\begin{figure*}
  \includegraphics[width=\textwidth]{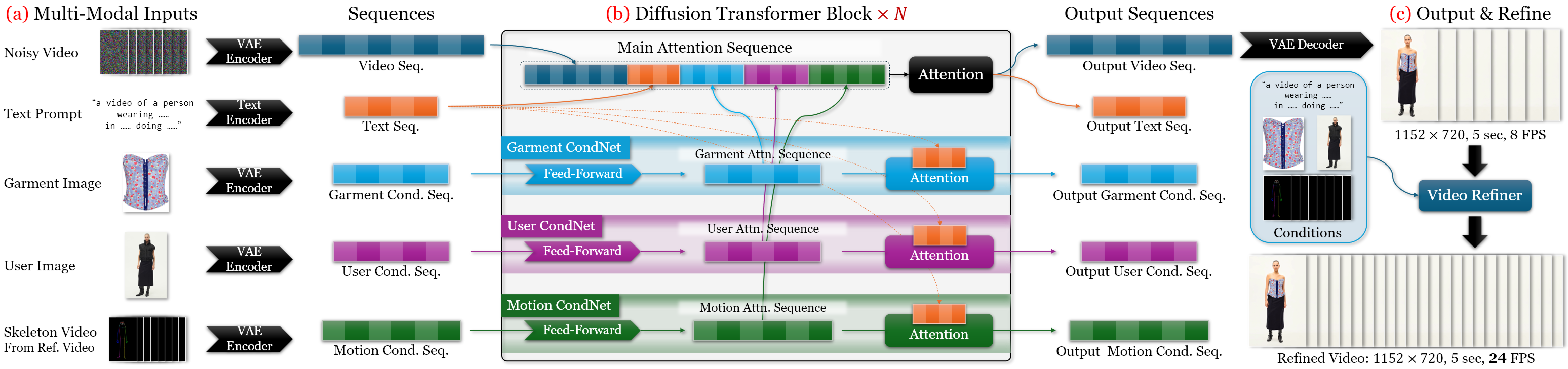}
  \vspace{-3mm}
  \caption{\fancyname supports multi-modal conditioning through our unified CondNets architecture based on attention mechanism. In a generation, the main diffusion model generates an 8-FPS video, and then the refiner model upsamples it to 24-FPS while removing artifacts.}
 
  \label{fig:method}
  \vspace{-3mm}
\end{figure*}

We propose \themodel, a novel video diffusion framework for virtual try-on, as illustrated in \figdot\ref{fig:method}. Given a user image, a garment or garment set to try on, a reference motion video, and an optional text prompt, \themodel synthesizes realistic videos of the user performing the indicated motion while wearing the desired garments. As shown in \figdot\ref{fig:method}\textcolor{red}{-(a)}, all inputs are encoded into token sequences and fed into a unified diffusion backbone. To support rich multi-modal conditioning, we introduce CondNets (\figdot\ref{fig:method}\textcolor{red}{-(b)}), modular attention-based adapters that inject garment, user, and motion cues into the diffusion process. A dedicated video refiner (\figdot\ref{fig:method}\textcolor{red}{-(c)}), fine-tuned from our main \themodel model, further upsamples the initial 8-FPS output to 24 FPS, while enhancing visual quality and removing artifacts.

To enable supervised training despite limited paired video data, we construct synthetic triplets, which eliminates the need for intermediates like ``agnostic masks'' or ``Dense Poses'' used in existing works, bridging the gap between training and inference formats. During optimization, we employ two key strategies: (1) a garment warm-up training phase based on curriculum learning, which guides the model to quickly learn garment registration—a core capability for try-on; and (2) a multi-stage progressive training scheme that gradually increases resolution and condition complexity, improving training stability and convergence. Together, these designs allow \themodel to deliver high-fidelity and controllable try-on results under diverse settings.

%% file: sec/4_expr_v2.tex
\section{Experiments}

\begin{figure}
\centering
  
  \includegraphics[width=1\linewidth]{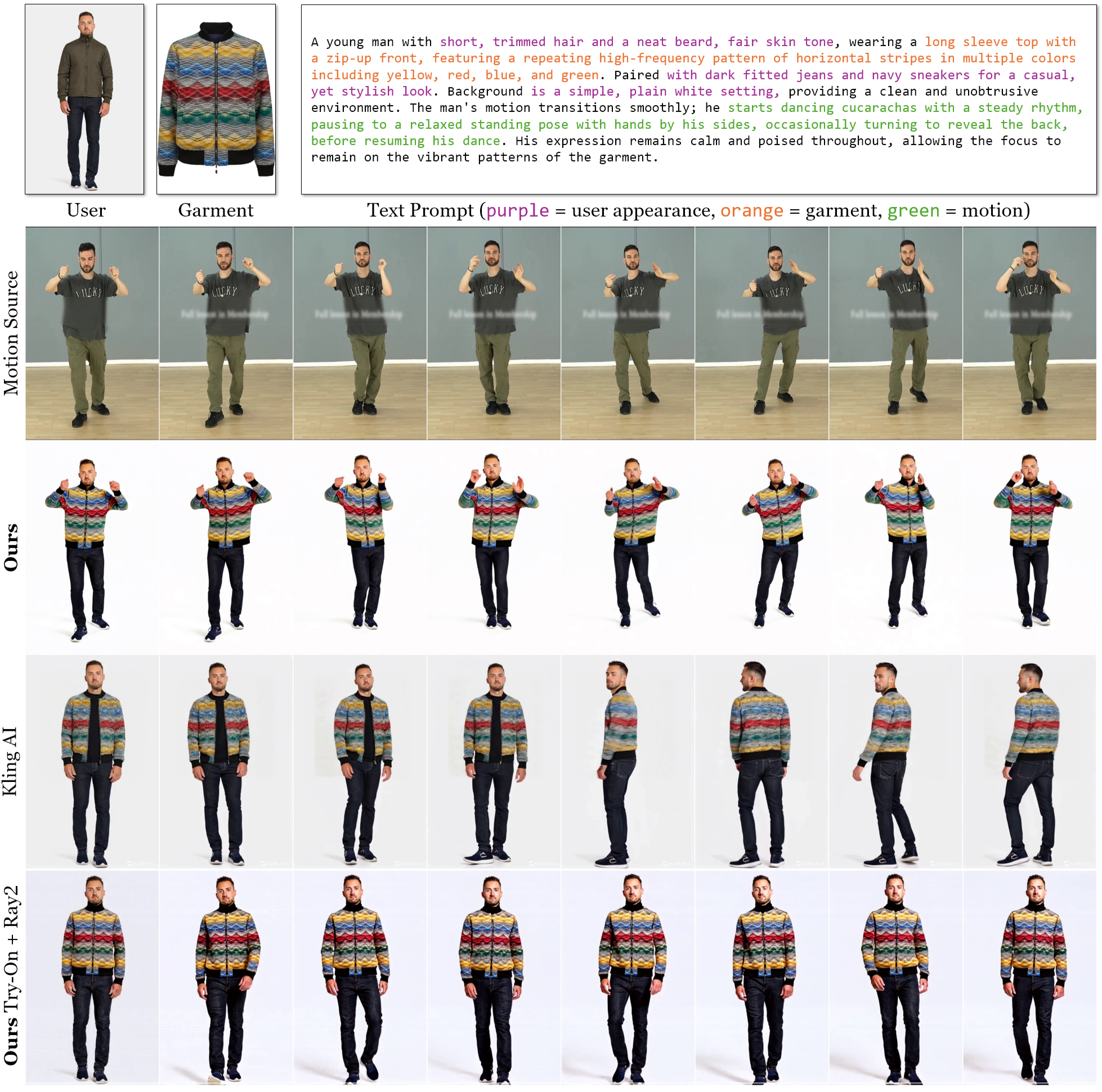}
  \vspace{-7mm}
  \caption{\fancyname allows a user to {\bf\color{lightblue} dress} up themselves with a desired garment, and perform the desired {\bf\color{red} dance}. It is difficult to express these moves by text which makes the generation of motion by Kling \cite{kling} and Ray2 \cite{ray2} quite challenging.  }
  \label{fig:expr:gallery:dance}
  \vspace{-6.5mm}
\end{figure}

\begin{figure*}
\centering
  \includegraphics[width=1\textwidth]{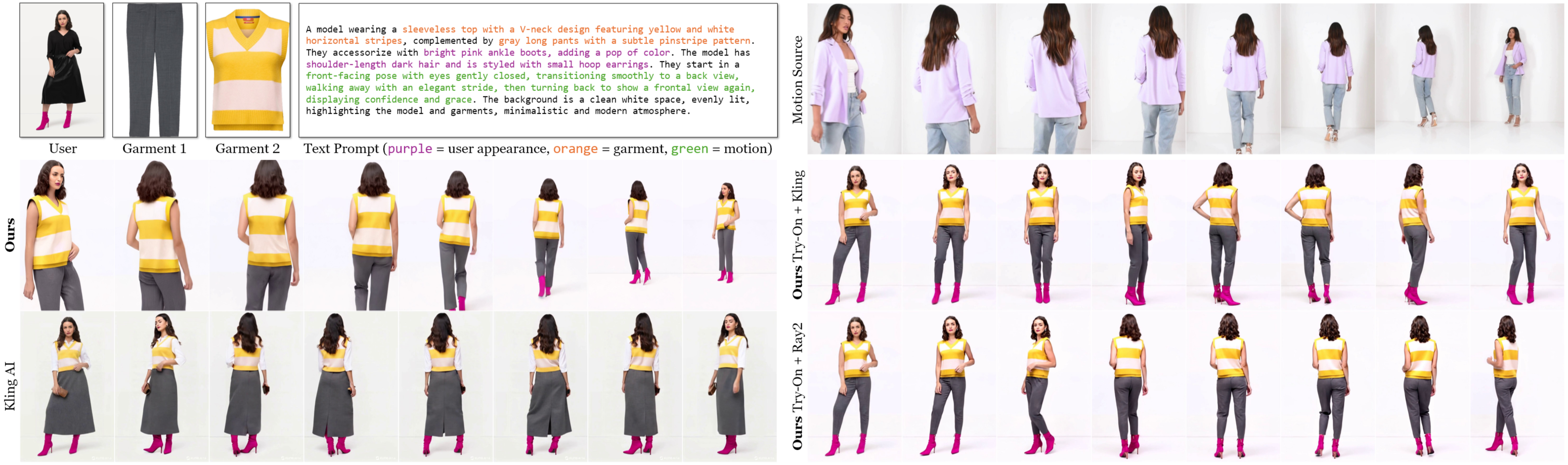}
  \vspace{-7mm}
  \caption{\fancyname supports simultaneously trying on a set of top and bottom garments, while correctly understanding and representing both garments without explicit labeling. On the contrary, Kling AI \cite{kling} misrepresents the trousers as a skirt. }
  \label{fig:expr:gallery:multigar}
  
\end{figure*}

\begin{figure*}
\centering
  \includegraphics[width=0.7\textwidth]{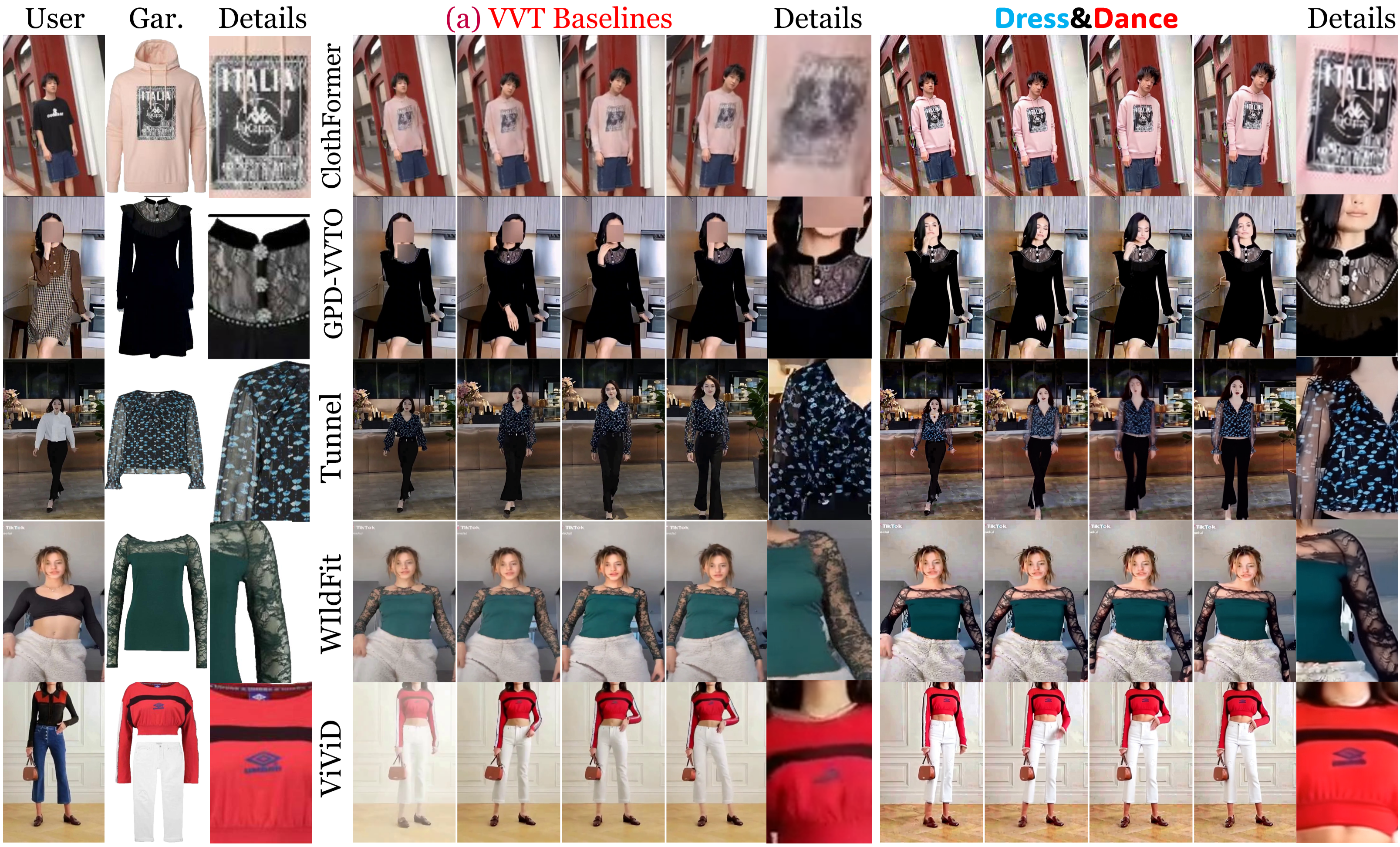}
  
  \caption{\fancyname significantly outperforms existing video virtual try-on methods, with much more detailed and precise textures and better support of transparent garments. }
  \label{fig:expr:gallery:vvt}
  \vspace{-6mm}
\end{figure*}

\begin{table*}[t!]
\centering

\scalebox{0.7}{\begin{tabular}{cc|l@{ }l|cccc}
 \hline\hline
 \multicolumn{2}{c|}{Method Property} &  \multicolumn{2}{c|}{\multirow{2}{*}{Method}} & \multicolumn{4}{c}{Evaluation with Ground Truth on Captured Dataset}\\
\cline{1-2} \cline{5-8}
Ours & Commercial  & & & PSNR$\uparrow$ & SSIM$\uparrow$ & $\text{LPIPS}_\text{VGG}\downarrow$ & $\text{LPIPS}_\text{AlexNet}\downarrow$\\
\hline

& & TPD & + CogVideoX I2V & 14.47 & 0.8305 & 0.2840 & 0.2461\\
& & OOTDiffusion&  + CogVideoX I2V & 14.68 & 0.8282 & 0.2779  & 0.2457\\
& & ML-VTON & + CogVideoX I2V & 14.49 & 0.8270 & 0.2961 & 0.2520 \\
\checkmark& & \fancyname Image Try-On & + CogVideoX I2V & 17.26 & 0.8515 & 0.2812 & 0.1635 \\
\hline
&\checkmark & Kling Image Try-On & + Kling Video 1.6 & \underline{17.33} & 0.8651 & \textbf{0.2296} & 0.1683 \\
\checkmark& \checkmark& \fancyname Image Try-On&  + Kling Video  1.6 & 15.21 & 0.8297 & 0.2835 & 0.2066\\
\checkmark& \checkmark& \fancyname Image Try-On&  + Ray2   & 15.16 & 0.8290 & 0.2938 & 0.1990 \\
\hline 
\checkmark&& \fancyname, Direct Train &  & 17.14 & \underline{0.8678} & 0.2854 & \underline{0.1338} \\
 \checkmark&& \fancyname &  & \textbf{22.41} & \textbf{0.9038} & \underline{0.2382} & \textbf{0.0624} \\
 \hline\hline
\end{tabular}}
\vspace{-3mm}

\caption{In the quantitative evaluations, \themodel significantly outperforms the open-sourced baselines under most of the metrics, while achieving comparable or even better quality metrics than commercial models Kling AI \cite{kling} and Ray2 \cite{ray2}. The \textbf{bold} numbers mark the best for each metric and the \textit{underlined} numbers mark the second best. }
\label{tab:exp:quan}
\vspace{-6mm}
\end{table*}

We study the performance of \fancyname using three different try-on modes (\figdot \ref{fig:modes}): (1) single garment mode allows the user to try on the garment shown in a single flat garment image, (2) ``a set of garments'' or multiple garment mode allows simultaneous try-on of all the garments provided as flat garment images, and (3) garment transfer mode allows to transfer a garment from an existing image through segmentation. 

\noindent\textbf{Datasets.} We curated two video datasets for training and evaluation. {\bf (1)} Internet video dataset is constructed by crawling publicly available fashion videos with paired flat garment images, which contains around $80K$ garment-video pairs. {\bf (2)} Our captured video dataset is constructed by hiring 183 human models to record the try-on videos for various sets of garments. Each model records the try-on videos for around 100 different sets of garments, so we can construct unpaired multi-garment try-on data with ground truth, by cross-matching the frames of different garments of the same person. We also collected an image dataset on the Internet with around 4M pairs of garment images paired for hybrid training. We divided each dataset into training and test subsets and performed training on the combination of all training subsets. For evaluation, we either sample garments and models from the evaluation subsets or use other garments, human images, and motion reference videos from the Internet.

\noindent\textbf{Baselines.} As there are no publicly available methods that support virtual try-on video generation, we compose the baseline methods using an ``image try-on + image animation'' pipeline: we first apply virtual try-on to the user image with a state-of-the-art image try-on method and then animate it according to the prompt with a video generation method. We consider two different types of baselines: 
{\bf (1) open-source models.} To perform single image try-on, we use state-of-the-art image try-on methods, TPD \cite{TPD} and OOTDiffusion \cite{ootdiffusion}. We also use ``ML-VTON'', an engineered version of GP-VTON~\cite{gpvton} that combines the original warping module with the HR-VTON~\cite{hrvton} try-on module. 
For comparison, we also provide the single image try-on generated by \themodel as a single-frame video. We then animate these try-on images with CogVideoX 1.5 I2V \cite{cogvideox}, one of the best open-source image-to-video (I2V) generation models available with text prompt descriptions. We cannot use other methods, including Stable Video Diffusion \cite{svd} and I2VGen-XL \cite{i2vgen}, as they do not support portrait videos well.
{\bf (2) commercial models.} We also compare with two state-of-the-art commercial methods for video generation: Kling Video 1.6 \cite{kling} and Ray2 \cite{ray2}. We once again perform image try-on with our \themodel on the user image and use these models to animate according to the text prompt. Kling AI's platform also provides an image try-on model. We also use it to perform a try-on for the subsequent video generation.
{\bf (3) video try-on baselines.} We compare with ViViD \cite{vivid}, WildFit \cite{wildvidfit}, Tunnel Try-On \cite{tunnel}, GPD-VVTO \cite{gpdvvto}, and ClothFormer \cite{clothformer} in ``single-garment'' setting, and Fashion-VDM \cite{fashionvdm} in the aligned ``garment transfer'' try-on task. Due to the lack of publicly available code, we compare our \themodel with the tasks presented on their website.

\subsection{Results and Analysis}

\noindent\textbf{Qualitative Comparison.}  We show the qualitative results in Figs\onedot \ref{fig:expr:gallery:singlegar}, \ref{fig:expr:gallery:seggar}, and \ref{fig:expr:gallery:multigar}. \figdot \ref{fig:expr:gallery:singlegar} show the try-on results in the single garment mode. More results are on our \svat{}. In the user image of \figdot \ref{fig:expr:gallery:singlegar}, the hand of the user covers the top-right corner of the garment. Therefore, any baseline, including Kling \cite{kling} and Ray2 \cite{ray2}, that first apply try-on to the user image and then animate it with another model, failed to generate the correct top-right garment pattern.  The information contained in the occluded garment is unable to be recovered, as the animation model does not consider the garment image. 
On the contrary, \themodel sees the garment image during the generation of the entire video and generates correct and consistent video try-on results. On our \svat{}, we evaluate our model with challenging dancing motions with two different dances, which are difficult to describe in text. This leads to the failure of all the baselines. Using the skeleton video extracted from the dancing video, our \themodel is able to generate an accurate and smooth dancing motion with the correct try-on results.

The results of the multiple garment mode are shown in \figdot \ref{fig:expr:gallery:multigar}. \themodel is able to perform the virtual try-on of both top and bottom garments simultaneously, \emph{without} any explicit information indicating the garment's type (\eg, which one is the top or the bottom). Regardless of the types and the orders of garments to try on, \themodel consistently generates high-quality try-on results. Kling AI, though officially supports multi-garment try-on, incorrectly tries on the trousers as a dress. 

In \figdot \ref{fig:expr:gallery:vvt}, we compare our model with multiple video try-on baselines in single garment mode; in \figdot \ref{fig:expr:gallery:seggar}, we compare our model with Fashion-VDM \cite{fashionvdm} in garment transfer mode. \themodel generates the videos at a higher resolution ($1152\times 720$), which better preserves the details and textures of the garments, and superior quality in rendering semi-transparent garments, while any other baseline's results are limited to a lower $512\times 384$ resolution with blurred and gloomy textures. 

\begin{table*}[t!]
\centering
\setlength{\tabcolsep}{4pt}

\scalebox{0.7}{\begin{tabular}{cc|l@{ }l|ccccccc}
 \hline\hline
  \multicolumn{2}{c|}{Method Property} &  \multicolumn{2}{c|}{\multirow{2}{*}{Method}} &  \multicolumn{7}{c}{Evaluation of Try-On and Visual Quality} \\
\cline{1-2} \cline{5-11}
Ours & Commercial  & & & $\text{GPT}_\text{Try-On}\uparrow$ &  $\text{GPT}_\text{User}\uparrow$  & $\text{GPT}_\text{Motion}\uparrow$ & $\text{GPT}_\text{Visual}\uparrow$ & $\text{GPT}_\text{Overall}\uparrow$ & $\text{FID}_\text{Internet}\downarrow$ & $\text{FID}_\text{Captured}\downarrow$\\
\hline

& & TPD & + CogVideoX I2V & 69.67& 73.98  & 68.15 & 65.45 & 68.64 &1146&753 \\
& & OOTDiffusion&  + CogVideoX I2V & 70.57  & 76.41 & 68.71 & 70.78 & 70.76  &1089&884 \\
& & ML-VTON & + CogVideoX I2V & 69.69  & 76.62 & 68.50 & 69.95 & 70.50 & 1185&739\\
\checkmark& & \fancyname Image Try-On & + CogVideoX I2V  & 85.05  & 87.54 & 75.68 & 76.88 & 80.71 &1109 &760 \\
\hline
&\checkmark & Kling Image Try-On & + Kling Video 1.6   & 80.10 & \textbf{89.97} & 85.48  & \textit{84.70} & 84.38 & \textbf{982} &\textbf{655}\\
\checkmark& \checkmark& \fancyname Image Try-On&  + Kling Video  1.6& \textit{86.85} &  \textit{89.77} & \textbf{82.59} & \textbf{84.94} & \textbf{85.85} & \textit{1008} &700 \\
\checkmark& \checkmark& \fancyname Image Try-On&  + Ray2    & 86.79 & 88.99 & 79.31 & 83.48 & 84.18 & 1094 &735 \\
\hline 
\checkmark&& \fancyname, Direct Train &   & 79.48 &  78.47 & 72.00 & 71.24 & 74.85 &1073 & 788 \\
 \checkmark&& \fancyname &   & \textbf{87.41} & \underline{88.89}  & \underline{\textit{80.35}} & \underline{84.48} & \underline{\textit{84.95}}  & \underline{1055} & \underline{\textit{691}} \\
 \hline\hline
\end{tabular}}
\caption{Our \themodel significantly outperforms all the baselines in garment fidelity in virtual try-on ($\text{GPT}_\text{Try-On}$), showing superior try-on capability, while achieving highly comparable or even better visual quality in other metrics to powerful commercial baselines Kling Video 1.6 \cite{kling} and Ray2 \cite{ray2}, which outperforms all open-sourced baselines with large margin. }
\label{tab:exp:quan2}
\end{table*}

\noindent\textbf{Quantitative Comparison.} We show the quantitative comparison in \tabdot \ref{tab:exp:quan}, where we conduct experiments on tasks constructed from our captured dataset, and compare the generated video with the ground truth using PSNR, SSIM, and LPIPS~\cite{lpips}. \themodel achieves results better than most methods. It also remains competitive to commercial models, Kling Video 1.6 \cite{kling} and Ray2 \cite{ray2}. 
For our Internet dataset, it is challenging to evaluate the try-on quality and visual quality, due to so many degrees of freedom in the generation. Inspired by VQAScore \cite{vqascore}, we leverage GPT \cite{gpt}'s strong capability in vision-language reasoning to grade the generated videos from the following aspects: garment try-on fidelity and quality ($\text{GPT}_\text{Try-On}$),  user appearance fidelity ($\text{GPT}_\text{User}$), human and garment motion quality ($\text{GPT}_\text{Motion}$), visual quality ($\text{GPT}_\text{Visual}$), and finally the overall quality considering all the aspects above ($\text{GPT}_\text{Overall}$). We provide detailed instructions and rubrics to guide the GPT to grade, while re-grading each video $40$ times and taking the average to avoid randomness and ensure fairness. The results are shown in \tabdot\ref{tab:exp:quan2}.

\underline{Virtual Try-On Capability.} As shown in the ``$\text{GPT}_\text{Try-On}$'' row, our \themodel significantly outperforms all the baseline models in garment fidelity and try-on quality. Also, using our \themodel for image try-on in baseline's pipeline consistently improves the try-on quality with large margins,  compared to the one using other image try-on methods, including Kling AI \cite{kling}. This shows that our \themodel has superior capability in virtual try-on.

\underline{Visual Quality.} In all other metrics, our \themodel achieves highly comparable results as the commercial models. Notably, as the commercial models are trained with much more video data and have the flexibility to generate motions only constrained by the text prompt, they are intrinsically easier to achieve high scores, but unable to perfectly present the indicated motion from the reference video. Even in this case, our \themodel still achieves very similar scores in all of the metrics, while the baseline, which achieves most of the best results, combines our try-on capability with Kling's animation.

\noindent\textbf{Ablation Study.} We study the effectiveness of our training strategy, including the garment warm-up and multi-stage training. We define the ``Direct Training (DT)'' variant for \themodel, which is directly trained with our final-stage inputs and outputs with full resolution. We compare full \themodel with the DT-variant in \tabdot \ref{tab:exp:quan} and \svat{3:14}. Without the garment warm-up training and multi-stage progressive training, the model is unable to faithfully preserve the contents of both user and garment images, even after extended training, resulting in low quantitative metrics. We observe that the training strategy is crucial for the convergence and final performance of the model.

%% file: sec/5_conclusion.tex
\section{Conclusion}
We introduce a novel video diffusion framework, \fancyname, that enables both garment try-on and temporally consistent motion generation. As the first work to achieve high-resolution ($1152\times 720$) video virtual try-on, our framework creates high-quality videos showing the user wearing a target garment, with motion guided by an example video. Our approach tackles several key challenges, including preserving the likeness of the user and garment, and generating complex motion with minimal error propagation. By utilizing a unified conditioning network with cross-attention, we effectively handle heterogeneous inputs, improving garment registration, and supporting various garment capture methods. Additionally, our data-efficient training strategy, coupled with synthetic triplet data generation and a multi-stage progressive approach, enables the generation of high-resolution videos with reduced artifacts.